\documentclass[journal]{IEEEtai}
\usepackage{subcaption}
\usepackage[colorlinks,urlcolor=blue,linkcolor=blue,citecolor=blue]{hyperref}
\usepackage{amsmath}
\usepackage{makecell}
\usepackage{amssymb}
\usepackage[x11names]{xcolor}
\usepackage{array}
\usepackage{multirow}
\usepackage{color}
\usepackage[dvipsnames, table]{xcolor}
\usepackage{graphicx}
\usepackage{diagbox}
\usepackage{pifont}
\usepackage{float}
\usepackage{placeins}
\usepackage{color}
\usepackage[normalem]{ulem} 

\begin{document}

\title{ASBench: Image Anomalies Synthesis Benchmark for Anomaly Detection} 

\author{Qunyi Zhang, Songan Zhang, Jiaqi Liu, Jinbao Wang, \textit{Member, IEEE}, Xiaoning Lei, Guoyang Xie, \textit{Member, IEEE}, Guannan Jiang, \textit{Member, IEEE}, Zhichao Lu, \textit{Member, IEEE}
\thanks{Received 10 October 2025; revised 9 January 2026; accepted 27 March 2026. This work was supported by the National Natural Science Foundation of
China under Grant 62576218. (Qunyi Zhang, Songan Zhang and Jiaqi Liu contributed equally
to this work. Corresponding authors: Guoyang Xie; Guannan Jiang.)}
\thanks{Qunyi Zhang and Songan Zhang are with Global Institute of Future Technology, Shanghai jJiao Tong University. (Email: zqyeleven@sjtu.edu.cn, songanz@sjtu.edu.cn)}
\thanks{Jinbao Wang is with School of Artificial Intelligence and also with the National Engineering Laboratory for Big Data System Computing Technology, Shenzhen University, Shenzhen 518060, China. (Email: wangjb@szu.edu.cn)}
\thanks{Xiaoning Lei, Guoyang Xie and Guannan Jiang are with Contemporary Amperex Technology Co.,Ltd. (Email: leixn01@outlook.com, guoyang.xie@ieee.org, jianggn@catl.com)}
\thanks{Jiaqi Liu and Zhichao Lu are with Department of Computer Science, City University of HongKong. (Email: liu\_jiaqi\_@outlook.com, zhichao.lu@cityu.edu.hk)}
}


\maketitle

\begin{abstract}
Anomaly detection plays a pivotal role in manufacturing quality control, yet its application is constrained by limited abnormal samples and high manual annotation costs. While anomaly synthesis offers a promising solution, existing studies predominantly treat anomaly synthesis as an auxiliary component within anomaly detection frameworks, lacking systematic evaluation of anomaly synthesis algorithms. Current research also overlooks crucial factors specific to anomaly synthesis, such as decoupling its impact from detection, quantitative analysis of synthetic data and adaptability across different scenarios. To address these limitations, we propose ASBench, the first comprehensive benchmarking framework dedicated to evaluating anomaly synthesis methods. Our framework introduces four critical evaluation dimensions: (i) the generalization performance across different datasets and pipelines (ii) the ratio of synthetic to real data (iii) the correlation between intrinsic metrics of synthesis images and anomaly detection performance metrics, and (iv) strategies for hybrid anomaly synthesis methods. Through extensive experiments, ASBench not only reveals limitations in current anomaly synthesis methods but also provides actionable insights for future research directions in anomaly synthesis. Code is available at https://github.com/M-3LAB/ASBench.
\end{abstract}

\begin{IEEEImpStatement}
Industrial image anomaly detection is a highly popular field in both academia and industry. However, academic research on industrial image anomaly detection primarily focuses on unsupervised anomaly detection, whereas industrial applications often directly employ supervised training methods to develop the required models. Anomaly synthesis can effectively bridge the gap between academia and industry by transforming unsupervised anomaly detection into supervised model training. Our paper presents the first comprehensive benchmark for analyzing anomaly synthesis in industrial images. Through a detailed examination of various anomaly synthesis methods, we aim to identify the currently optimal synthesis method and narrow the gap between industrial practices and academic research.
\end{IEEEImpStatement}


\section{Introduction}

\IEEEPARstart{A}{nomaly} detection has emerged as a critical technology in modern manufacturing quality control and healthcare monitoring, playing an important role in ensuring product reliability and enhancing diagnostic accuracy~\cite{10494062}~\cite{9971462}. Despite its significance, the practical deployment of anomaly detection systems is often constrained by two major challenges: the scarcity of abnormal samples and the prohibitively high costs associated with manual annotation. These limitations significantly impede the performance and scalability of the detection models. To address these issues, anomaly synthesis has gained traction as a promising solution, offering the capability to generate representative abnormal samples. This approach not only alleviates the data scarcity problem but also substantially reduces the reliance on costly manual annotation efforts. Jan and Christian (2022)~\cite{diers2023survey} summarize the anomaly synthesis methods.  The significant potential has been demonstrated in various domains, including surface defect detection and medical image analysis. Anomaly synthesis can be effectively integrated with real anomaly data or directly applied to enhance detection systems.

While anomaly synthesis holds transformative potential for advancing anomaly detection capabilities, current anomaly synthesis methods remain largely confinded to auxiliary roles within anomaly detection frameworks. This paradigm neglects rigorous benchmarking and detailed exploration of the methodological nuances and performance characteristics unique to anomaly synthesis algorithms. The absence of systematic evaluation frameworks creates critical gaps in understanding algorithmic strengths, operational boundaries and domain-specific adaptability-limitations. To address this critical gap, our reasearch focuses specifically on anomaly synthesis algorithms, with the primary objectives of establishing a systematic evaluation framework and conducting rigorous performance analysis. The proposed framework aims to provide scientific foundations for anomaly detection tasks, offering theoretical guidance for the selection and optimization of synthesis methods in practical applications.

Existing benchmark studies in this field primarily focus on anomaly detection tasks, yet fail to address critical challenges specific to anomaly synthesis. The current research landscape exhibits the following limitations:
\begin{enumerate}

    \item There has been no systematic evaluation of different anomaly synthesis methods across various datasets, thereby lacking guidance in practical applications.

    \item Since the detection and generation components of the model are not decoupled, it is impossible to assess the magnitude of the impact of anomaly synthesis on the performance of anomaly detection models.
    
    \item Instead of direclty conducting quantitative analysis on synthesized abnormal images, the effectiveness of anomaly synthesis methods is only indirectly reflected through the metrics of anomaly detection models
    
    \item An analysis of how the data proportion of synthetic anomaly samples influences anomaly detection tasks is absent.
    
    \item Current anomaly detection frameworks predominantly employ single anomaly synthesis algorithms, neglecting potential benefits from hybrid application of multiple synthesis strategies.

\end{enumerate}

\begin{table*}[h!]
\centering
\scriptsize
\resizebox{\textwidth}{!}{
\begin{tabular}{l|cccccccc}
\hline
Off-the-shelf Work \& Ours & IM-IAD~\cite{xie2024iad} & ADer~\cite{zhang2024ader} & MMAD~\cite{jiang2024mmad} & MulSen-AD~\cite{li2024multi} & RAD~\cite{cheng2024radcomprehensivedatasetbenchmarking} & Real-IAD~\cite{wang2024real} & PAD~\cite{zhou2023paddatasetbenchmarkposeagnostic} & \textcolor{blue}{ASBench} \\
\hline
Contains Anomaly Synthesis Method & \checkmark & \ding{55} & \ding{55} & \ding{55} & \ding{55} & \checkmark & \checkmark & \checkmark \\

Decoupling of Synthesis and Detection & \ding{55} & \ding{55} & \ding{55} & \ding{55} & \ding{55} & \ding{55} & \ding{55} & \checkmark \\

Multiple Datasets & 7 & 11 & 4 & 1 & 1 & 1 & 1 & 5 \\

Different Abnormal Data Ratio & \ding{55} & \ding{55} & \ding{55} & \ding{55} & \ding{55} & \ding{55} & \ding{55} & \checkmark \\

Different Pipeline Combination & \ding{55} & \ding{55} & \ding{55} & \ding{55} & \ding{55} & \ding{55} & \ding{55} & \checkmark \\
\hline
\end{tabular}
}
\caption{Comparison of different anomaly detection benchmarks. Our ASBench is the first to provide a decoupled analysis of the individual components in anomaly synthesis methods.}
\label{tab:anomaly_synthesis_benchmark}
\end{table*}

    To address these limitation, we propose ASBench, a comprehensive benchmark to evaluate the anomaly synthesis from four perspectives: 
\begin{enumerate}
\item \textbf{Cross Dataset and Detection Methods Comparison}: Cross-dataset generalization capability and detection model compatibility evaluation of individual synthesis methods.

\item \textbf{Data Ratio Impact}: Investigate the impact of different sample proportions of synthesis anomalies on the performance of anomaly detection models.

\item \textbf{Metric Correlation}: Correlation analysis between intrinsic quality metrics of the anomaly images generated and downstream detection performance metrics.

\item \textbf{Hybrid Strategies}: Evaluate the effects of mixing abnormal data samples generated by multiple anomaly synthesis algorithms.
\end{enumerate}

\begin{figure*}[htbp]
    \centering
    \includegraphics[width=1\textwidth]{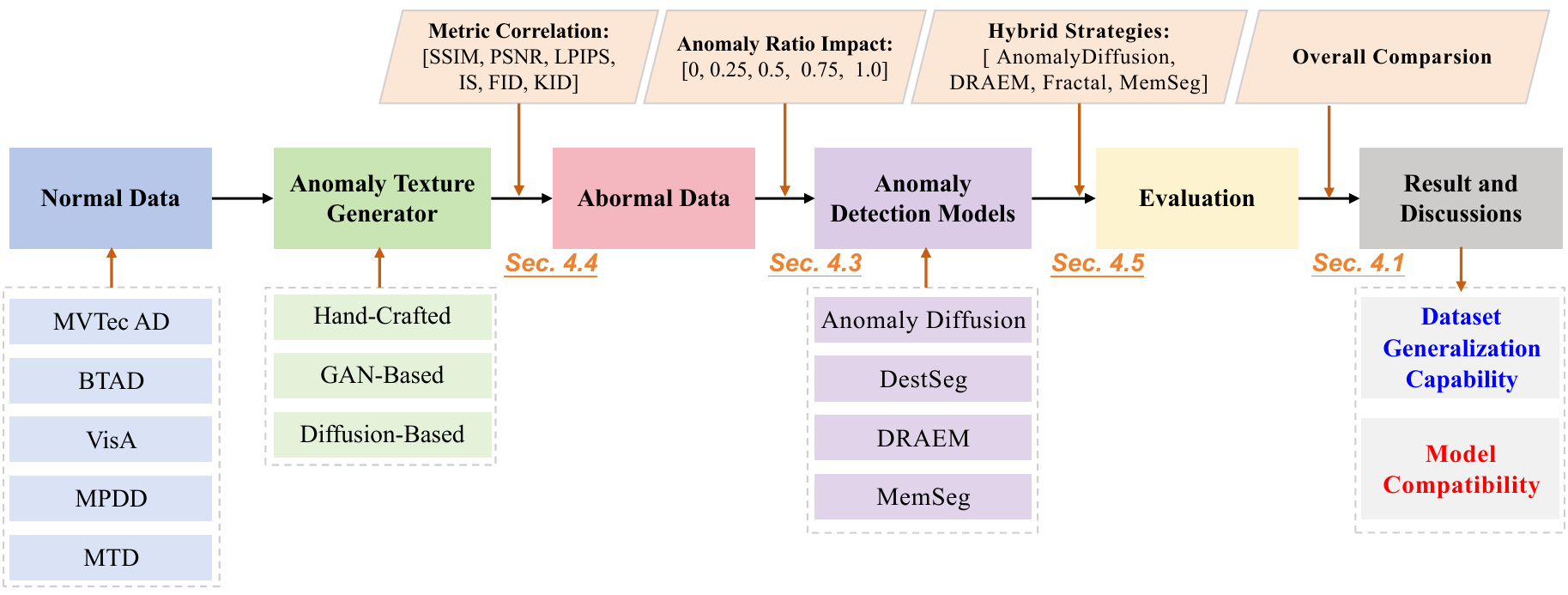} 
    \caption{The overall workflow of ASBench. We disentangle different stages in the anomaly synthesis and anomaly detection pipeline, and discuss the impact of variables at each stage on anomaly detection in separate subsections of Section IV.}
    \label{fig:overall_workflow}
\end{figure*}

As shown in Table~\ref{tab:anomaly_synthesis_benchmark}, compared to other benchmark studies in the field of anomaly detection, our ASBench stands out for its unique features in anomaly synthesis decoupling, adjustable anomaly sample ratios, and diverse pipeline configurations. The overall workflow of ASBench, along with the chapter structure detailing analyzes from various perspectives, is illustrated in Fig.~\ref{fig:overall_workflow}.


\textbf{Key Takeaways:} Through extensive experimentation, we have identified several key insights:
\begin{enumerate}
    \item No single anomaly synthesis method universally dominate across all datasets and algorithms.

    \item The sample proportions of generated anomalies exert no significant influence on the performance of anomaly detection models. 
    
    \item There is no correlation between the intrinsic metrics of generated images and detection performance.

    \item Combined usage of multiple anomaly synthesis can improve the accuracy of anomaly detection.
\end{enumerate}

Our main contributions are listed below.
\begin{enumerate}
\item \textbf{First Comprehensive Benchmark for Anomaly Synthesis}: To address the research gap in evaluation standards for anomaly generation, we propose ASBench, a standardized and unified benchmark designed for comprehensive investigation and experimental assessment of anomaly generation methodologies.
\item \textbf{Open-IAS Framework for Unified Evaluation}: We propose Open-IAS, a flexible framework integrating 12 anomaly synthesis methods, 4 detection pipelines, and 5 industrial datasets, resulting in a total of 19,680 data. providing standardized protocols and reproducible baselines to accelerate future research in anomaly synthesis.
\item \textbf{Critical Analysis and Future Directions}: Through ASBench, we perform granular comparisons across anomaly proportions, generation strategies, dataset adaptability, and evaluation metrics. This reveals limitations in existing methods and presents actionable research directions.
\end{enumerate}

 The remainder of this paper is structured as follows. Section II reviews related literature on anomaly detection pipelines and anomaly synthesis methods. Section III introduces the experimental settings of our proposed benchmark, ASBench. In Section IV, a comprehensive experimental evaluation is presented along with a discussion of the results from four perspectives. Finally, Section V concludes the paper.

\section{Related Work}
 
\subsection{Anomaly Detection}
With the release of the MVTec AD dataset~\cite{Bergmann_2019_CVPR}, the development of industrial image anomaly detection has shifted from a supervised paradigm to an unsupervised one~\cite{10970745,10510400}. In the unsupervised paradigm, the training set consists solely of normal images, while the test set includes both normal and labeled anomalous images. Research in unsupervised industrial image anomaly detection has gradually evolved into two main categories: feature embedding-based methods and reconstruction-based methods.

Feature embedding-based methods can be further divided into four subcategories. Specifically, (i) teacher-student distillation models typically comprise a pretrained teacher network and a trainable student network. During training, the knowledge of normal sample features extracted by the teacher network is distilled into the student network. During inference, discrepancies between features extracted by the student network for anomalous samples and those extracted by the teacher network facilitate anomaly detection. State-of-the-art results in this category have been achieved by methods such as RD4AD and RD++. 
(ii) One-class classification methods generate anomalies either at the image level or the feature level and then learn to classify anomalous images or features. Representative methods in this subcategory include CutPaste~\cite{li2021cutpaste} and SimpleNet~\cite{Liu2023SimpleNetAS}.
(iii) Mapping-based methods utilize pretrained models to extract image features, which are then mapped to a desired distribution using a feature mapping module. During testing, if the sample's features deviate from the expected distribution, the sample is deemed anomalous. Techniques in this category frequently employ normalizing flow modules to map features to a multivariate Gaussian distribution.
(iv) Memory-based methods use pretrained networks to extract features from training samples, which are subsequently sampled and stored in a memory bank. During testing, anomaly scores are computed by measuring the distance between the test sample's features and the features stored in the memory bank.

Reconstruction-based methods share a similar overall architecture. These methods typically involve self-supervised training, wherein normal images and artificially generated anomalous images are reconstructed into normal images. Anomaly localization is achieved during testing by comparing the differences between the reconstructed and original images. Autoencoders are the most commonly used reconstruction networks for anomaly detection, as seen in DRAEM~\cite{zavrtanik2021draem}, DSR~\cite{zavrtanik2022dsr}, and NSA~\cite{schluter2022natural}. While methods leveraging generative adversarial networks (GANs) as reconstruction networks are fewer, they have demonstrated outstanding performance, indicating untapped potential in this area. Recently, transformers, as a foundational model in computer vision, have also been employed as reconstruction networks and have shown impressive results in anomaly detection. Diffusion models, which are among the most popular generative models currently, have set new SOTA benchmarks in anomaly detection tasks with methods such as DDAD~\cite{mousakhan2024anomaly} and DiffusionAD~\cite{zhang2025diffusionad}.

The field of anomaly detection has seen excellent development in available datasets. MVTec AD~\cite{Bergmann_2019_CVPR}, as the first comprehensive dataset, covering multiple object and defect types, has become the widely recognized benchmark in the field. MTD~\cite{Huang2018SurfaceDS} is specifically designed for saliency detection in tile images, focusing on the identification of anomalies at a micro-scale. BTAD~\cite{Mishra_2021}  is a dataset from real industrial scenarios where images have relatively simple backgrounds, making the detection task more straightforward. MPPD~\cite{9631567} focuses on defects in metal painted parts during the manufacturing process, emphasizing detection capability under diverse viewing angles and complex conditions. The innovation of the VisA dataset~\cite{zou2022spot}  lies in its introduction of a multi-instance sample setting. These datasets align with the objectives of the anomaly detection tasks targeted by ASBench from various perspectives.

PAD~\cite{zhou2023pad} and Real-IAD~\cite{wang2024real} share a key characteristic with MPPD by focusing on multi-viewpoint imaging to closely approximate real industrial conditions. Comprehensive 3D datasets such as MVTec 3D-AD~\cite{bergmann2021mvtec3dad} and Real3D-AD~\cite{liu2023real3d} are primarily oriented towards anomaly detection and localization in 3D data, which does not fully align with ASBench's current focus on 2D image detection tasks.

\subsection{Anomaly Synthesis}
The development of anomaly generation techniques has evolved from simple image‑level manipulations to complex generative approaches.
Hand-crafted method is the mainstream research direction in the early stages of anomaly synthesis research. Initial methods primarily operated at the patch level for augmentation. Techniques such as CutOut~\cite{devries2017improvedregularizationconvolutionalneural} and RIAD~\cite{zavrtanik2021reconstruction} simulated anomalies by blacking out certain patches, while methods like CutPaste~\cite{li2021cutpaste} further replaced patches with textures from other regions to mimic anomalies.  However, the shapes of anomalies generated through these approaches often differed significantly from real anomalous shapes. Consequently, subsequent research gradually shifted toward using random noise to simulate anomalous regions. Representative methods include DRAEM~\cite{zavrtanik2021draem}, which employs Perlin noise~\cite{hart2001perlin} and external texture datasets to generate anomalous textures, and FractalAD~\cite{xia2024fractalad}, which utilizes fractal noise. Building upon these approaches, MemSeg\cite{yang2023memseg} constrained anomalies to appear only in foreground regions, whereas NSA~\cite{schluter2022natural} filled anomalous areas with the object’s own texture via Poisson image editing.

With the advancement of generative models, methods based on generative models have gradually become mainstream in anomaly synthesis, as they produce textures that are closer to real anomalies and offer greater diversity. Early methods such as DFMGAN~\cite{duan2023few} and AdaBLDM~\cite{li2024novel} utilized GANs~\cite{salimans2016improved} and latent diffusion models to generate defective images, thereby improving anomaly generation capabilities under few‑shot conditions. In recent years, diffusion models~\cite{ho2020denoising} have been introduced to the anomaly synthesis task. Methods such as AnomalyDiffusion~\cite{hu2024anomalydiffusion} enable precise control over the spatial distribution and visual appearance of anomalies, capable of generating more realistic anomalous images. RealNet~\cite{zhang2024realnet} combined adaptive diffusion models with feature selection modules to enhance anomaly detection performance.

Based on the distinct characteristics of existing anomaly synthesis methods, we have selected patch augmentation-based and random noise-based methods from the Hand-crafted synthesis category, as well as GANs-based and diffusion-based methods from the generation-model-based category for our benchmark.

\section{ASBench}

\subsection{Problem Definition}
 The proposed ASBench systematically addresses three critical dimensions in various anomaly synthesis evaluation: cross-dataset adaptability, detection algorithm compatibility, and anomaly ratio sensitivity.  
The focus of ASBench is on existing anomaly detection models and related anomaly synthesis techniques. This research first evaluates the generated images of anomaly synthesis methods independently by decoupling the anomaly detection and synthesis components. Subsequently, it combines existing anomaly synthesis methods with different anomaly detection models, summarizing the performance across various industrial datasets. Additionally, the interactions between multiple anomaly synthesis methods are evaluated to identify the most effective and realistic anomaly generation strategies, in order to explore the optimal approach for anomaly detection tasks in industrial environments. 

The experimental framework of this study can be formally summarized by the following relationship:
$$
Trained \ Model = A \otimes ( B, C, D),
$$
where $A$ represents the anomaly detection algorithm, $B$ denotes the anomaly synthesis method, $C$ corresponds to the industrial dataset and $D$ specifies the proportion of synthetic abnormal samples in the training data. The $\otimes$ operator represents composability between components. This formulation establishes a systematic approach to evaluate the interdependencies between critical components.

\subsection{Implementation Details}
The research conducts systematic comparisons across 5 industrial anomaly detection datasets by integrating 12 anomaly synthesis approaches with 4 detection models. The overall workflow is shown in Fig.~\ref{fig:overall_workflow}, the anomaly synthesis and detection processes are decoupled into different stages.

Table~\ref{tab:anomaly_synthesis_methods_intro} lists 12 anomaly synthesis methods used in ASBench (marked in purple). The criteria for selecting methods to be implemented for ASBench are that different shapes of the abnormal region and whether model training is required. 

\begin{table*}[t]
\centering
\resizebox{1\textwidth}{!}{%
\begin{tabular}{cc|cllllll}
\hline
\multicolumn{2}{c|}{Paradigm}                                                        & \multicolumn{7}{c}{Methods}                                          \\ \hline
\multicolumn{1}{c|}{\multirow{2}{*}{Hand-Crafted}}           & Patch Augmentation    & \multicolumn{7}{c}{\textcolor{blue}{CutOut}~\cite{devries2017improvedregularizationconvolutionalneural}, \textcolor{blue}{CutPaste}, \textcolor{blue}{CutPaste\_Scar}~\cite{li2021cutpaste}, \textcolor{blue}{FPI}~\cite{Tan_2022}, PII~\cite{tan2021detecting}, RIAD~\cite{zavrtanik2021reconstruction}} \\ \cline{2-9} 
\multicolumn{1}{c|}{}                                        & Random Shape Noise    & \multicolumn{7}{c}{\textcolor{blue}{DestSeg}~\cite{Zhang2022DeSTSegSG}, \textcolor{blue}{DRAEM}~\cite{zavrtanik2021draem}, \textcolor{blue}{FractalAD}~\cite{xia2024fractalad}, \textcolor{blue}{MemSeg}~\cite{yang2023memseg}, \textcolor{blue}{NSA}~\cite{schluter2022natural}}            \\ \hline
\multicolumn{1}{c|}{\multirow{2}{*}{Generative-Model-Based}} & GAN-Based             & \multicolumn{7}{c}{\textcolor{blue}{DFMGAN}~\cite{duan2023few}, Con-GAN~\cite{du2022new}, Defect-GAN~\cite{zhang2021defect}}                      \\ \cline{2-9} 
\multicolumn{1}{c|}{}                                        & Diffusion-Model-Based & \multicolumn{7}{c}{\textcolor{blue}{RealNet}~\cite{zhang2024realnet}, \textcolor{blue}{AnomalyDiffusion}~\cite{hu2024anomalydiffusion}, DefectDiffu~\cite{shi2024few}, AdaBLDM~\cite{li2024novel}} \\ \hline
\end{tabular}
}
\caption{Representative anomaly synthesis algorithms for ASBench. The blue ones indicate our re-implemented method.}
\label{tab:anomaly_synthesis_methods_intro}
\end{table*}
Regarding the selection of anomaly detection pipelines, our research incorporates four representative architectures: \text{DR\AE M}, DestSeg, MemSeg, and AnomalyDiffusion. \text{DR\AE M} employs pixel-level composition and segmentation for anomaly identification, while DestSeg leverages feature distillation and contrastive learning to effectively capture subtle discrepancies in complex textures. In contrast to \text{DR\AE M}'s pixel-level fusion strategy, MemSeg operates through feature space segmentation, demonstrating particular efficacy in scenarios requiring distinct anomaly separation within feature representations. AnomalyDiffusion adopts a novel paradigm by decoupling abnormal appearance and structural information, directing the model's attention to minute anomalous patterns. These algorithms exhibit complementary strengths in segmentation mechanisms, anomaly saliency processing, and feature comparison strategies, contributing to the development of a robust detection system capable of addressing diverse anomaly characteristics.

For conducting thorough ablation studies, we utilize five publicly available datasets, MVTec AD~\cite{Bergmann_2019_CVPR}, BTAD~\cite{Mishra_2021}, VisA~\cite{zou2022spot}, MPDD~\cite{9631567}, and MTD~\cite{Huang2018SurfaceDS}. Table~\ref{tab:dataset_intro} offers a detailed summary of these datasets, encompassing the number of samples (covering both training and test sets, with normal and abnormal samples), the total number of classes, image resolution, and the primary characteristics of each dataset. All datasets include pixel-level annotations. 

\begin{table*}
\centering
\resizebox{0.8\textwidth}{!}{%
\begin{tabular}{c|c|c|c|c|c|c|c}
\hline \multirow{3}{*}{Datasets}& \multicolumn{3}{c|}{Sample Number} & \multirow{3}{*}{Classes} & \multicolumn{2}{l|}{Image Resolution} & \multirow{3}{*}{Main Feature} \\
\cline{2-4} \cline{6-7} & Train Set & \multicolumn{2}{c|}{Test Set} & & \multirow{2}{*}{Min} & \multirow{2}{*}{Max} & \\
\cline{2-4} & Normal & Normal & Abnormal & & & & \\ 
\hline MVTec AD~\cite{Bergmann_2019_CVPR}& 3629 & 467 & 1258 & 15 & 700 & 1024 & Basic benchmark standard \\
\hline BTAD~\cite{Mishra_2021}& 1799 & 451 & 290 & 3 & 600 & 1600 & Real-world manufacturing \\
\hline VisA~\cite{zou2022spot} & 8659 & 962 & 1200 & 12 & 960 & 1562 & Multi-instance IAD \\
\hline MPDD~\cite{9631567} & 888 & 176 & 282 & 6 & 1024 & 1024 & Multi-view Anomalies \\
\hline MTD~\cite{Huang2018SurfaceDS} & 902 & 50 & 392 & 1 & 113 & 491 & Micro-scale anomalies \\
\hline
\end{tabular}}
\caption{Statistics of the selected datasets with diverse features used in ASBench.}
\label{tab:dataset_intro}
\end{table*}

Regarding evaluation metrics, we employ Area Under the Receiver Operating Characteristics (AU-ROC/AUC), Area Under Precision-Recall (AUPR/AP), and \textcolor{blue}{Per-Region Overlap (PRO)}~\cite{bergmann2021mvtecIJCV} to evaluate the abilities of anomaly localization. These five metrics collectively address both anomaly detection and segmentation performance at image and pixel granularities, ensuring comprehensive and precise evaluation.

For experimental configurations, all parameter settings strictly adhere to the default specifications outlined in the detection models' publicly available source codes, maintaining methodological consistency and reproducibility.

\section{Result and Discussions}
This section explores current anomaly synthesis methods and discusses the essential components of the proposed uniform settings. Each subsection outlines the experimental methodologies employed, presents a detailed analysis of the results, and identifies unresolved challenges along with potential avenues for future research.
\subsection{Overall Comparison}

\begin{table*}[htbp]
\centering
\resizebox{\textwidth}{!}{%
\begin{tabular}{c|c|c|c|c|c|c|c|c|c|c}
\hline Dataset & \multicolumn{2}{c|}{MVTEC} & \multicolumn{2}{c|}{BTAD} & \multicolumn{2}{c|}{VisA} & \multicolumn{2}{c|}{MPDD} & \multicolumn{2}{c}{MTD} \\
\hline
 Metric & \makecell{AUC \\ Image} & \makecell{AP \\ Pixel} & \makecell{AUC \\ Image} & \makecell{AP \\ Pixel} & \makecell{AUC \\ Image} & \makecell{AP \\ Pixel} & \makecell{AUC \\ Image} & \makecell{AP \\ Pixel} & \makecell{AUC \\ Image} & \makecell{AP \\ Pixel}\\

\hline AnomalyDiffusion & \cellcolor{yellow!10}0.9796 & \cellcolor{blue!10}0.7599 & \cellcolor{red!10}0.9481 & \cellcolor{green!10}\textbf{0.6748} & \cellcolor{red!10}0.8813 & \cellcolor{red!10}0.4179 & \cellcolor{red!10}\textbf{0.9883} & \cellcolor{green!10}\textbf{0.4640} & \cellcolor{blue!10}\textbf{0.9732} & \cellcolor{red!10}\textbf{0.6801} \\
\hline CutOut & \cellcolor{blue!10}0.8799 & \cellcolor{red!10}0.2644 & \cellcolor{green!10}0.8796 & \cellcolor{red!10}0.1630 & \cellcolor{yellow!10}0.7959 & \cellcolor{red!10}0.1832 & \cellcolor{red!10}0.9218 & \cellcolor{red!10}0.2018 & \cellcolor{blue!10}0.7449 & \cellcolor{red!10}0.1150 \\
\hline CutPaste & \cellcolor{red!10}0.9539 & \cellcolor{red!10}0.5659 & \cellcolor{blue!10}\textbf{0.9643} & \cellcolor{green!10}0.4159 & \cellcolor{red!10}0.8558 & \cellcolor{red!10}0.3576 & \cellcolor{red!10}0.9181 & \cellcolor{yellow!10}0.1688 & \cellcolor{red!10}0.8870 & \cellcolor{blue!10}0.4892 \\
\hline CutPaste Scar & \cellcolor{red!10}0.9417 & \cellcolor{red!10}0.3851 & \cellcolor{red!10}0.8848 & \cellcolor{red!10}0.2078 & \cellcolor{red!10}0.8529 & \cellcolor{red!10}0.2542 & \cellcolor{blue!10}0.8918 & \cellcolor{yellow!10}0.1969 & \cellcolor{red!10}0.8428 & \cellcolor{red!10}0.1128 \\
\hline DestSeg & \cellcolor{red!10}\textbf{0.9892} & \cellcolor{red!10}\textbf{0.7948} & \cellcolor{blue!10}0.9445 & \cellcolor{red!10}0.5010 & \cellcolor{blue!10}\textbf{0.9508} & \cellcolor{yellow!10}0.3727 & \cellcolor{red!10}0.9749 & \cellcolor{blue!10}0.3401 & \cellcolor{red!10}0.9476 & \cellcolor{red!10}0.3754 \\
\hline DFMGAN & \cellcolor{red!10}0.9393 & \cellcolor{red!10}0.6027 & \cellcolor{red!10}0.9397 & \cellcolor{green!10}0.3355 & \cellcolor{blue!10}0.8208 & \cellcolor{yellow!10}0.2879 & \cellcolor{green!10}0.8832 & \cellcolor{red!10}0.2696 & \cellcolor{green!10}0.6902 & \cellcolor{yellow!10}0.0807 \\
\hline DRAEM & \cellcolor{red!10}0.9889 & \cellcolor{red!10}0.7810 & \cellcolor{red!10}0.9500 & \cellcolor{red!10}0.5049 & \cellcolor{blue!10}0.9362 & \cellcolor{red!10}\textbf{0.4933} & \cellcolor{red!10}0.9819 & \cellcolor{blue!10}0.3774 & \cellcolor{red!10}0.9470 & \cellcolor{red!10}0.3565 \\
\hline FPI & \cellcolor{blue!10}0.9557 & \cellcolor{red!10}0.6527 & \cellcolor{blue!10}0.9366 & \cellcolor{green!10}0.3508 & \cellcolor{blue!10}0.9338 & \cellcolor{red!10}0.4030 & \cellcolor{blue!10}0.9300 & \cellcolor{yellow!10}0.2487 & \cellcolor{blue!10}0.8405 & \cellcolor{red!10}0.2200 \\
\hline Fractal & \cellcolor{red!10}0.9858 & \cellcolor{red!10}0.7327 & \cellcolor{blue!10}0.9442 & \cellcolor{red!10}0.5621 & \cellcolor{red!10}0.9348 & \cellcolor{red!10}0.4866 & \cellcolor{red!10}0.9598 & \cellcolor{green!10}0.4193 & \cellcolor{red!10}0.9090 & \cellcolor{red!10}0.2710 \\
\hline MemSeg & \cellcolor{red!10}0.9841 & \cellcolor{red!10}0.7515 & \cellcolor{red!10}0.9469 & \cellcolor{red!10}0.5836 & \cellcolor{green!10}0.9025 & \cellcolor{red!10}0.2568 & \cellcolor{green!10}0.9257 & \cellcolor{red!10}0.3070 & \cellcolor{red!10}0.9322 & \cellcolor{red!10}0.3129 \\
\hline NSA & \cellcolor{blue!10}0.9813 & \cellcolor{blue!10}0.6957 & \cellcolor{green!10}0.9541 & \cellcolor{green!10}0.5175 & \cellcolor{blue!10}0.9210 & \cellcolor{red!10}0.4822 & \cellcolor{blue!10}0.9625 & \cellcolor{yellow!10}0.2703 & \cellcolor{blue!10}0.8674 & \cellcolor{red!10}0.1927 \\
\hline RealNet & \cellcolor{red!10}0.9815 & \cellcolor{red!10}0.7130 & \cellcolor{green!10}0.9235 & \cellcolor{green!10}0.2788 & \cellcolor{green!10}0.8601 & \cellcolor{red!10}0.2358 & \cellcolor{red!10}0.9395 & \cellcolor{red!10}0.2939 & \cellcolor{red!10}0.9191 & \cellcolor{red!10}0.2207 \\
\hline
\end{tabular}}
\caption{Optimal algorithm performance of 12 synthesis methods on 5 datasets. Bolded data represents the optimal results among the twelve methods. Different colors correspond to different detection models: red represents DestSeg, blue represents DRAEM, green represents MemSeg, and yellow represents AnomalyDiffusion.}
\label{tab:Comparative_Analysis_of_12_Synthesis_Methods}
\end{table*}

\subsubsection{Comparative Experimental Framework}
\begin{figure*}[htbp]
    \centering
    \begin{minipage}{0.9\textwidth}
        \centering
        \includegraphics[width=1\linewidth]{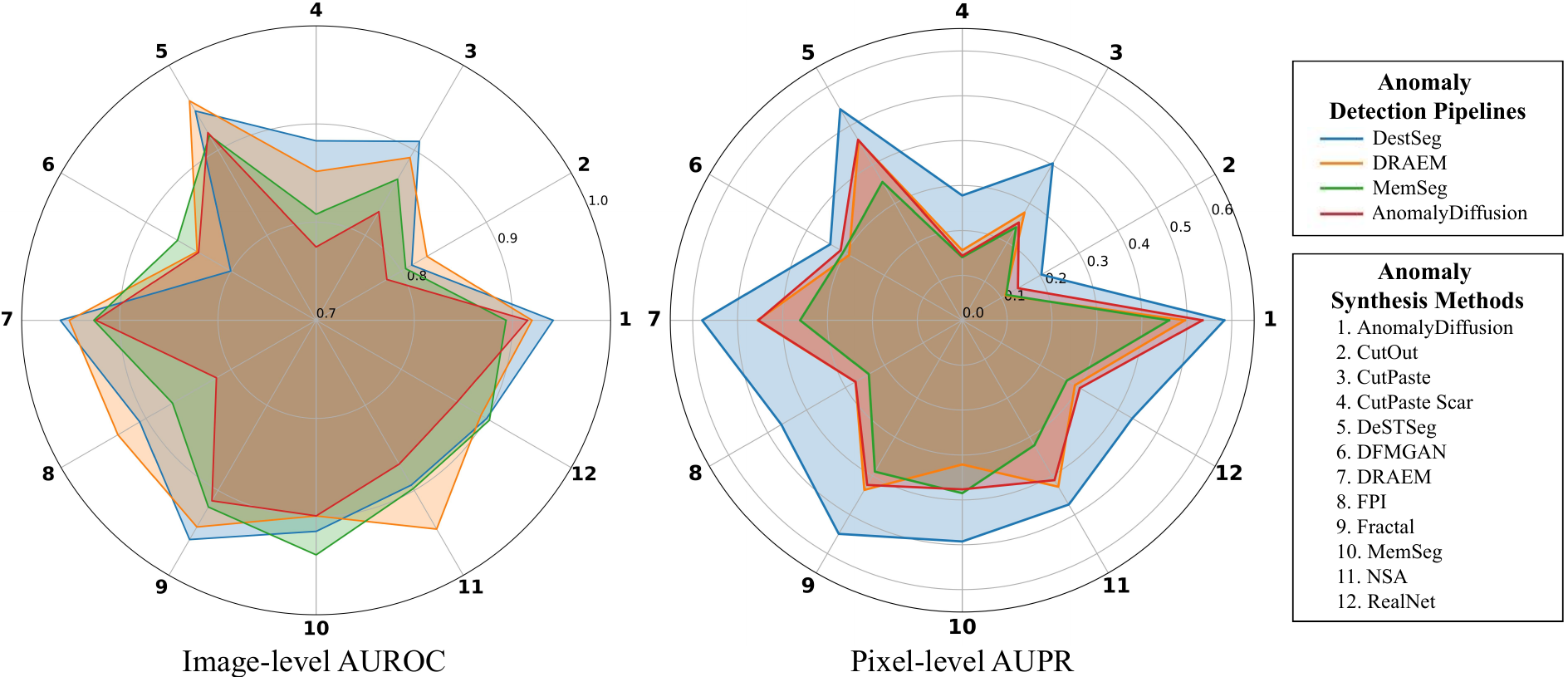}
        \label{fig:Sec1_All_Datasets_AUROC}
    \end{minipage}
    
    \caption{Performance comparison of different anomaly synthesis methods across various detection pipelines and all datasets. For datasets, results are computed via weighted averaging, where weights correspond to the number of subclasses per dataset. Axes represent the anomaly synthesis methods, lines correspond to the detection pipelines, and vertices quantify the performance of these synthesis methods across different detection pipelines.}
    \label{fig:Sec1_All_Datasets}
\end{figure*}

We conducted a comprehensive comparison across 5 datasets, 12 synthesis methods, and 4 detection algorithms, and the  detailed results are presented in Table~\ref{tab:Comparative_Analysis_of_12_Synthesis_Methods}. To visualize the overall performance, we generated radar charts (Fig.~\ref{fig:Sec1_All_Datasets}) depicting AUC Image and AP Pixel aggregated from all datasets. Notably, for the overall evaluation, we adopted a weighted averaging approach using the number of subclasses within each dataset as weighting coefficients to account for inter-dataset categorical imbalances.

The weighted average formula is as follows:
\begin{equation}
\overline{M} = \sum_{i=1}^{5} w_i \cdot M_{i}, \quad w_i = \frac{N_i}{\sum_{j=1}^{5} N_j},
\label{eq:weighted_avg}
\end{equation}
where $M_{i}$ denotes the metric value on dataset $i$, $N_i$ is the number of subclasses in dataset $i$, and weights $w_i$ normalize subclass counts to mitigate categorical imbalance.

\begin{figure*}[htbp]
    \centering
    \begin{minipage}{1\textwidth}
        \centering
        \includegraphics[width=\linewidth]{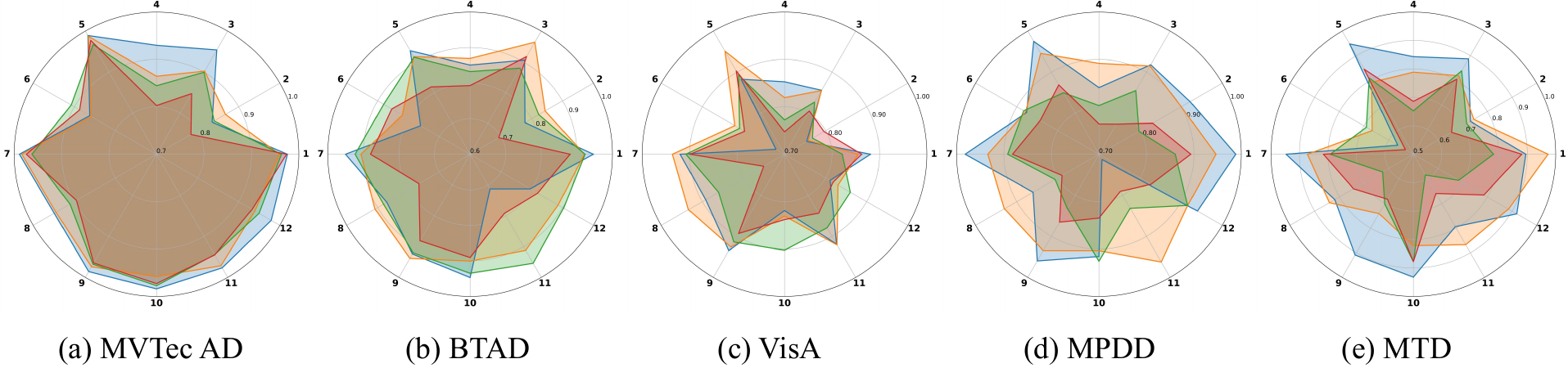}

        \label{fig:AUC_Image_Comparison_Radar_datasets}
    \end{minipage}
    
    \caption{Image-level AUROC performance comparison of different anomaly synthesis methods across various detection pipelines and different datasets.}
    \label{fig:comparison_radar_AUC_Image}
\end{figure*}

\begin{figure*}[htbp]
    \centering
    \begin{minipage}{1\textwidth}
        \centering
        \includegraphics[width=\linewidth]{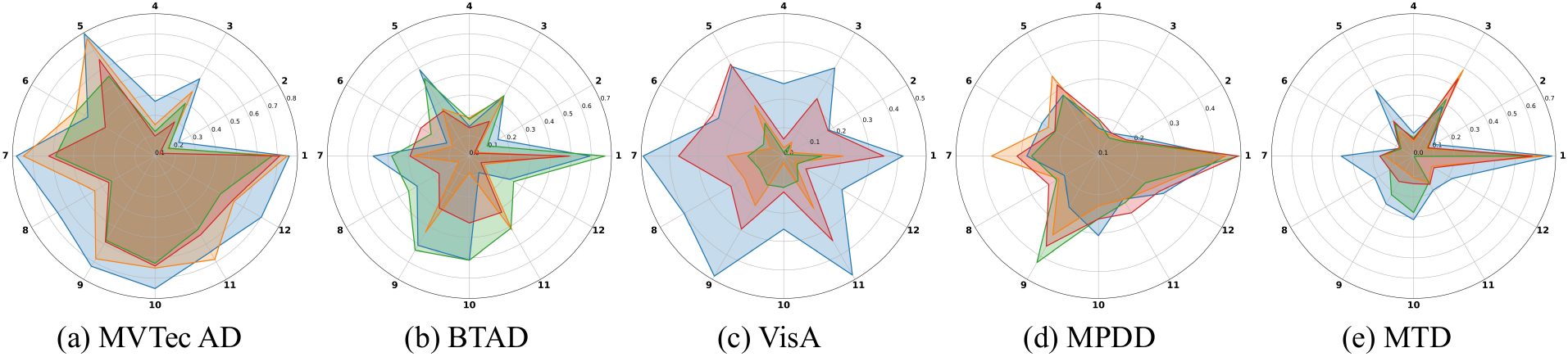}
    \end{minipage}
    
    \caption{Pixel-level AUPR performance comparison of different anomaly synthesis methods across various detection pipelines and different datasets.}
    \label{fig:comparison_radar_AP_Pixel}
\end{figure*}

Additionally, Fig.~\ref{fig:comparison_radar_AUC_Image} and \ref{fig:comparison_radar_AP_Pixel} present separate radar charts that illustrate AUROC Image and AUPR Pixel metrics for individual datasets, providing granular insights into the performance of methods across different data environments. 

\subsubsection{Analysis of Results}
The experimental results indicate that no single anomaly synthesis method demonstrates universal superiority.  Our investigation reveals that current anomaly synthesis methods and detection algorithms, predominantly validated on the MVTec AD dataset. Consequently, optimal performance on MVTec AD is achieved by employing the synthesis method originally proposed within the respective detection algorithms. However, when these approaches are extended to alternative datasets and integrated with different processing algorithms, the previously superior detection algorithms and synthesis methods fail to maintain their performance advantages, demonstrating significant performance degradation in cross-dataset applications.

As evidenced by Fig.~\ref{fig:comparison_radar_AUC_Image} and Table~\ref{tab:Comparative_Analysis_of_12_Synthesis_Methods}, significant performance variations are observed across different combinations of detection algorithm and synthesis method on various datasets. Regarding image-level anomaly classification, the four synthesis methods, DestSeg, DRAEM, Fractal, and AnomalyDiffusion exhibit generally stable performance across all four detection algorithms. With notable exceptions: CutPaste combined with DRAEM's algorithm achieves state-of-the-art results on BTAD, where textures are relatively simple and homogeneous. In contrast, the VisA dataset contains multi-instance objects, making it more suitable for methods like FPI and NSA that employ seamless image editing and blending, or Fractal-based techniques that generate diversified anomalies through synthetic fractal patterns. For datasets focusing on multi-view, fine-grained details (e.g., MPDD, MTD), generative models such as AnomalyDiffusion and RealNet perform better, as they excel at simulating detailed anomalies. For pixel-level anomaly localization, the AnomalyDiffusion detection pipelines demonstrates robust compatibility with most synthesis methods on VisA, MPDD, BTAD, and MTD datasets. However, on the MVTec AD dataset, synthesis strategies derived from DRAEM and DestSeg yield superior localization performance compared to AnomalyDiffusion.  

\begin{table*}[htbp]
\centering
\resizebox{0.9\textwidth}{!}{%
\begin{tabular}{c|c|c|c|c|c}
\hline
{Dataset} & MVTec AD & BTAD & VisA & MPDD & MTD \\
\hline
DestSeg($\mathcal{D}$) & DestSeg & DRAEM & Fractal & AnomalyDiffusion & DestSeg \\
\hline
DRAEM($\mathcal{D}$) & DestSeg & CutPaste & DestSeg & NSA & AnomalyDiffusion \\
\hline
MemSeg($\mathcal{D}$) & MemSeg & NSA & Fractal & MemSeg & MemSeg \\
\hline
AnomalyDiffusion($\mathcal{D}$) & AnomalyDiffusion & CutPaste & DestSeg & AnomalyDiffusion & AnomalyDiffusion \\
\hline
\end{tabular}}
\caption{Optimal anomaly synthesis methods across different algorithms and datasets based on image-level AUROC. The $\mathcal{D}$ represents detection pipeline.}
\label{tab:Optimal_ASMethods}
\end{table*}

\subsubsection{Implications and Future Challenges}
The experimental findings underscore necessity of strategic selection between detection pipelines and synthesis methods based on task-specific requirements.   Optimal performance requires dual adaptation: first, to dataset-specific characteristics (e.g., background complexity and positional variance), and second, to task priorities (classification versus localization).  This empirical evidence motivates pursuit of a universal anomaly synthesis methods capable of autonomously adapting to heterogeneous data distributions while maintaining cross-algorithm compatibility. This is a critical direction for advancing generalizable anomaly detection systems.

\subsection{Analysis of Anomaly Synthesis Methods Across Datasets and Detection Models}
\subsubsection{Comparative Experimental Framework}
\begin{figure*}[htbp]
    \centering
    \begin{subfigure}[b]{\textwidth}
        \centering
        \includegraphics[width=\linewidth]{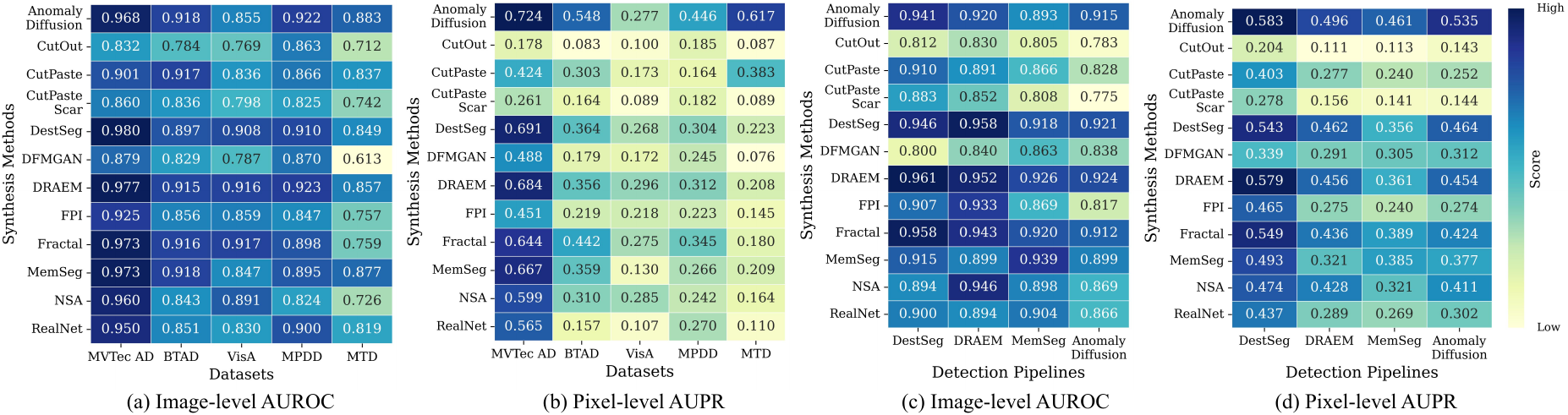}
        \label{fig:auroc_heatmap}
    \end{subfigure}
    \caption{Heatmaps of image-level AUROC and pixel-level AUPR performance for anomaly synthesis methods. The images (a) and (b) represent performance across different datasets. The images (c) and (d) represent performance across different detection pipelines. AnomalyDiffusion, DRAEM, and DestSeg establish a clear advantage in  anomaly synthesis effects over other methods.}
    \label{fig:Heatmap_4_detection_ppl}
\end{figure*}

Based on the experiments conducted across five datasets, twelve synthesis methods, and four detection models, we further analyze the following aspects.
First, the cross dataset and detection model performance analysis is performed.
We compute the average performance metrics for each synthesis method across detection pipelines. We construct the heatmap (Fig.~\ref{fig:Heatmap_4_detection_ppl}) with synthesis methods versus datasets to visually present performance variations across different dataset-method combinations.
By averaging the performance metrics across all five datasets, we generate another heatmap (Fig.~\ref{fig:Heatmap_4_detection_ppl}), which takes synthesis methods and detection models as axes. Fig.~\ref{fig:Heatmap_4_detection_ppl} primarily highlights the performance fluctuations of identical synthesis methods across different detection models.


\begin{figure}[htbp]
    \centering
    \includegraphics[width=1.0\columnwidth]{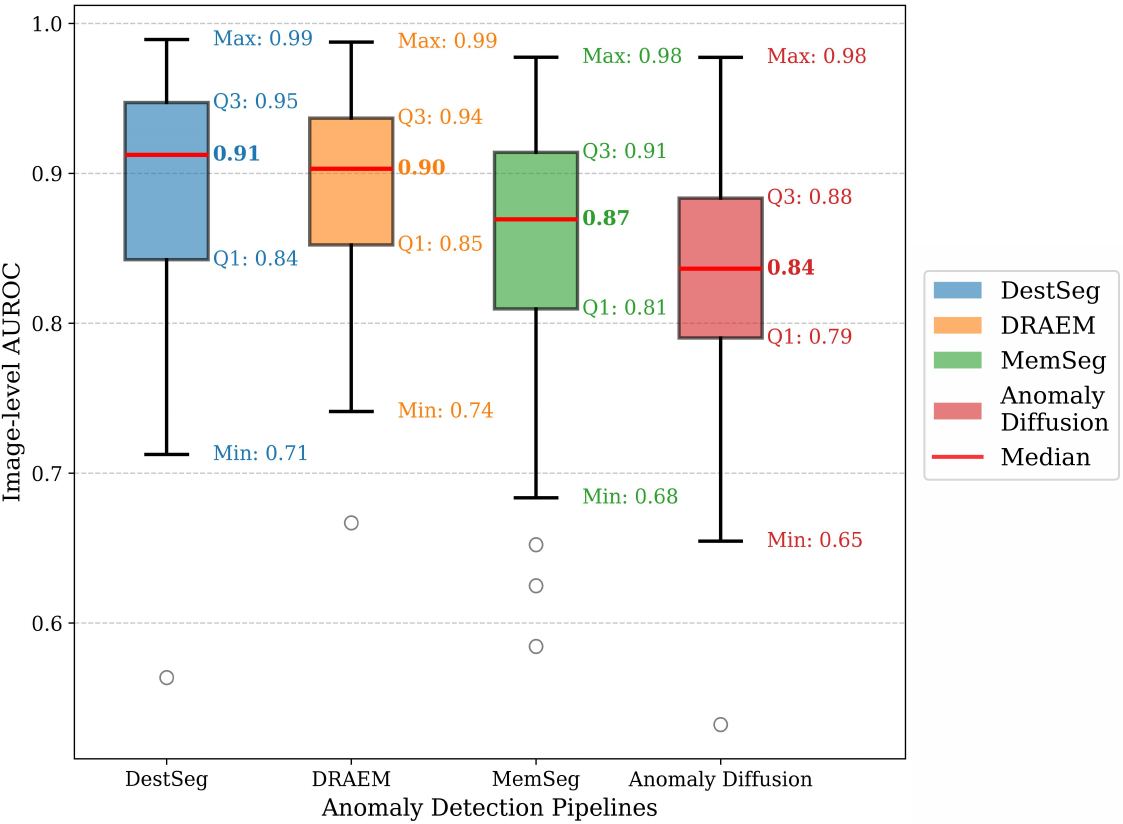} 
    \caption{Comparison of detection pipelines: (a) DestSeg, (b) DRAEM, (c) MemSeg, and (d) Anomaly Diffusion.}
    \label{fig:box_plots}
\end{figure}
Then, to analyze model robustness, we calculate the mean and variance of performance metrics for each synthesis method across different datasets and under different detection models. And we construct Fig.~\ref{fig:box_plots} to quantify the discrepancy in robustness among the methods. 

\subsubsection{Analysis of Results}
As shown in the Fig.~\ref{fig:Sec1_All_Datasets} and Fig.~\ref{fig:Heatmap_4_detection_ppl}, through comparative analysis of synthesis methods, we observe that approaches capable of generating complex and diverse anomalous regions significantly outperform handcrafted designs or those relying on real anomaly samples, and the potential advantages of real anomalies remaining underutilized. Handcrafted methods like such as CutPaste, CutOut, and FPI demonstrate inferior performance. Approaches leveraging real anomalies, including DFMGAN, RealNet, and AnomalyDiffusion, also do not exhibit significant advantages. The more effective methods, DRAEM, DestSeg, Fractal, and NSA, employ complex and diverse masks to increase anomaly diversity and mitigate overfitting risks. Defining anomalous regions through Fractal patterns or Perlin Noise proves more effective than random cropping or simulating real anomalies, as these techniques introduce greater stochasticity that improves model generalization capabilities.

Notably, MemSeg synthesis method restricts anomalies to object foregrounds, unlike DRAEM, to reduce background noise interference. However, this strategy yields no significant performance gains; in some cases,  MemSeg underperforms DRAEM. The excessive foreground concentration limits anomaly diversity, failing to cover potential background noise patterns and increasing false positives. Since background noise is typically subtle, models focusing solely on foregrounds overlook such anomalies, compromising overall detection effectiveness. Moreover, MemSeg's overly strict foreground constraints require dataset-specific parameter tuning for optimal results, diminishing its generalization capability and further impairing detection performance.

Fig.~\ref{fig:Heatmap_4_detection_ppl} reveals that DestSeg and DRAEM achieve the highest average performance among the evaluated approaches, with DRAEM exhibiting particularly low variance. DestSeg's strength stems from its two-stage training strategy: In the first stage, it aligns the student model’s features with those of the teacher model, adopting the distillation-based anomaly detection framework from prior unsupervised methods~\cite{bergmann2020uninformed,deng2022anomaly}. This allows knowledge transfer without requiring anomaly samples. The second stage introduces a segmentation model for anomaly localization while keeping the pre-trained teacher-student models fixed. Building on this foundation of first stage, DestSeg delivers robust performance across diverse synthetic anomalies. In addition, DRAEM demonstrates minimal sensitivity to anomaly data ratios and synthesis method variations. Its reconstruction-based architecture enables training under extreme conditions—even in anomaly-free scenarios, the model can localize anomalies by detecting discrepancies between input images and their reconstructions.

\subsubsection{Implications and Future Challenges}

Based on the preceding analysis, the optimal synthesis methods are influenced by detection models and dataset characteristics. These collective observations underscore that no single synthesis method is universally optimal across all evaluation scenarios, necessitating careful calibration among datasets, evaluation metrics, and downstream detection pipelines. The core challenge lies in designing highly generalizable anomaly synthesis approaches that balance diversity and authenticity in generated anomalies. Future work should explore more efficient utilization of real data while developing models robust to background noise and cross-dataset discrepancies.

\begin{figure}[htbp]
    \centering
    \includegraphics[width=1.0\columnwidth]{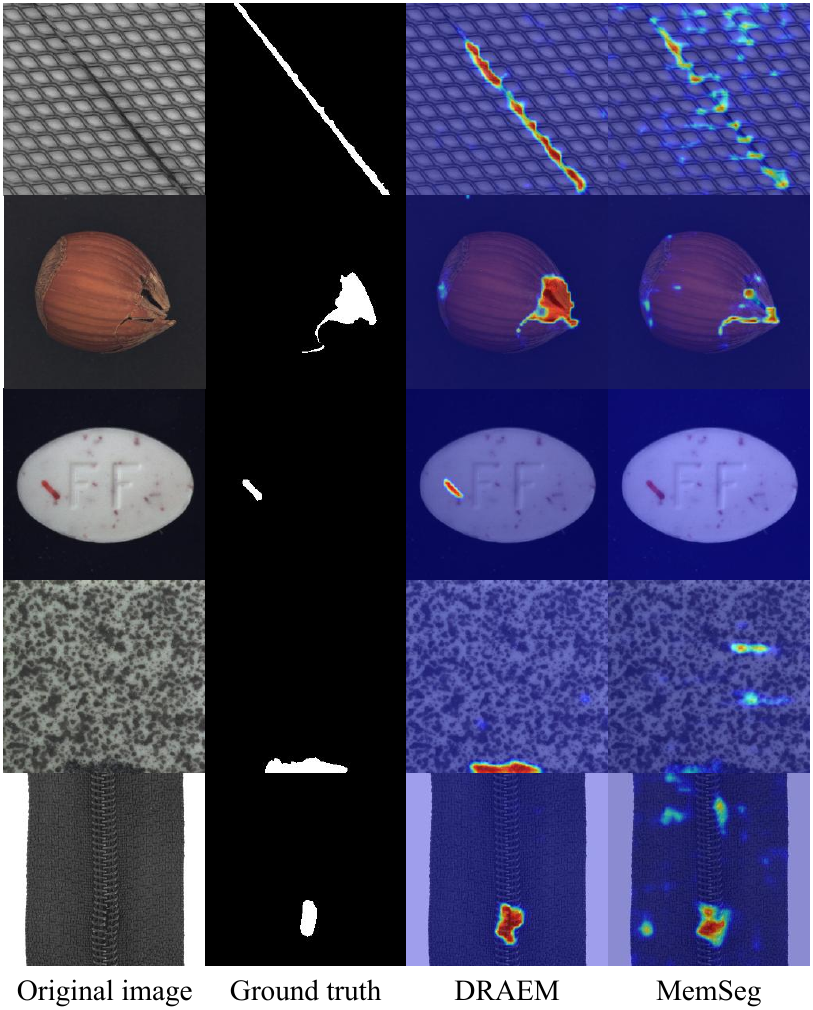}
    \caption{Visualization comparison between synthesis methods: DRAEM and MemSeg.}
    \label{fig:vis_draem_memseg}
\end{figure}

As shown in Fig.~\ref{fig:vis_draem_memseg}, the comparison between MemSeg and DRAEM synthesis methods reveals that anomaly synthesis methods must strike a balance between anomaly diversity and background complexity, as overemphasizing foreground objects can yield detrimental effects. Harmonizing anomaly diversity with background noise management emerges as the pivotal challenge for advancing anomaly synthesis and detection performance. For instance, adjust masks while preserving fidelity to real anomalies; or synergistically integrate authentic anomaly-derived masks with procedurally generated masks (e.g., Perlin/fractal noise) to enhance structural diversity.

Furthermore, regarding downstream detection pipelines, current stable pipelines universally incorporate U-Net for segmentation, underscoring its essential role in ensuring robustness. Future work should optimize U-Net for unsupervised industrial scenarios, particularly for long-tail distributions where anomalies are sparse, to fundamentally enhance detection generalization capabilities.

\subsection{Effect of Different Anomaly Ratios}

\subsubsection{Comparative Experimental Framework}
Considering the original implementations, the default anomaly-to-total ratios (i.e., the proportion of anomalous samples in the training set) were set to 0.5 and 1.0 across the four detection algorithms. For our experiments, we systematically varied this ratio in the training sets (0.25, 0.5, 0.75, and 1.0) on the MVTec AD and BTAD datasets.

This configuration enables dual analytical perspectives: (a) averaging across four detection models to evaluate how data ratios affect performance under different synthesis strategies, and (b) averaging across synthesis methods to examine how anomaly ratios impact individual model performance.  On the other hand, we investigate the relationship between synthesis method stability and performance by analyzing how the performance differential (between optimal and worst-case scenarios) varies with anomaly sample ratios. The model-averaged performance results and the performance differentials are recorded in Table~\ref{tab:Performance_Proportion_MVTec} and \ref{tab:Performance_Proportion_BTAD}. Based on tabulated mean values and performance gaps, we generate scatter plots and calculate the related correlation coefficients to analyze.

\begin{table*}[h!]
\centering
\begin{tabular}{c|c|c|c|c}
\hline
Dataset & DestSeg & DRAEM & MemSeg & AnomalyDiffusion  \\
\hline
Metric& AUROC / AUPR / PRO & AUROC / AUPR / PRO & AUROC / AUPR / PRO & AUROC / AUPR / PRO \\
\hline
MVTec AD & 0.5850 / 0.0362 / 0.1536 & 0.7425 / 0.0531 / 0.2324 & 0.7968 / 0.1440 / 0.5065 & 0.7985 / 0.0730 / 0.3416 \\
\hline
BTAD & 0.5421 / 0.0330 / 0.1572 & 0.7578 / 0.0572 / 0.2436 & 0.6277 / 0.0442 / 0.4859 & 0.8028 / 0.0361 / 0.2822 \\
\hline
VisA & 0.5276 / 0.0071 / 0.1850 & 0.7462 / 0.0046 / 0.3265 & 0.6852 / 0.0082 / 0.4862 & 0.5666 / 0.0500 / 0.6029 \\
\hline
MPDD & 0.3735 / 0.0345 / 0.1874 & 0.7808 / 0.0756 / 0.2718 & 0.7832 / 0.1411 / 0.4326 & 0.7642 / 0.1676 / 0.6418 \\
\hline
MTD & 0.6345 / 0.0820 / 0.1646 & 0.7401 / 0.0742 / 0.2150 & 0.5620 / 0.0743 / 0.4753 & 0.4814 / 0.0776 / 0.2921 \\
\hline
\end{tabular}
\caption{Performance comparison (AUROC/AUPR/PRO) of various methods under anomaly-free conditions on different datasets.}
\label{tab:combined_scores_simple}
\end{table*}

\subsubsection{Analysis of Results}
\begin{figure*}[htbp]
\centering
\includegraphics[width=0.9\linewidth]{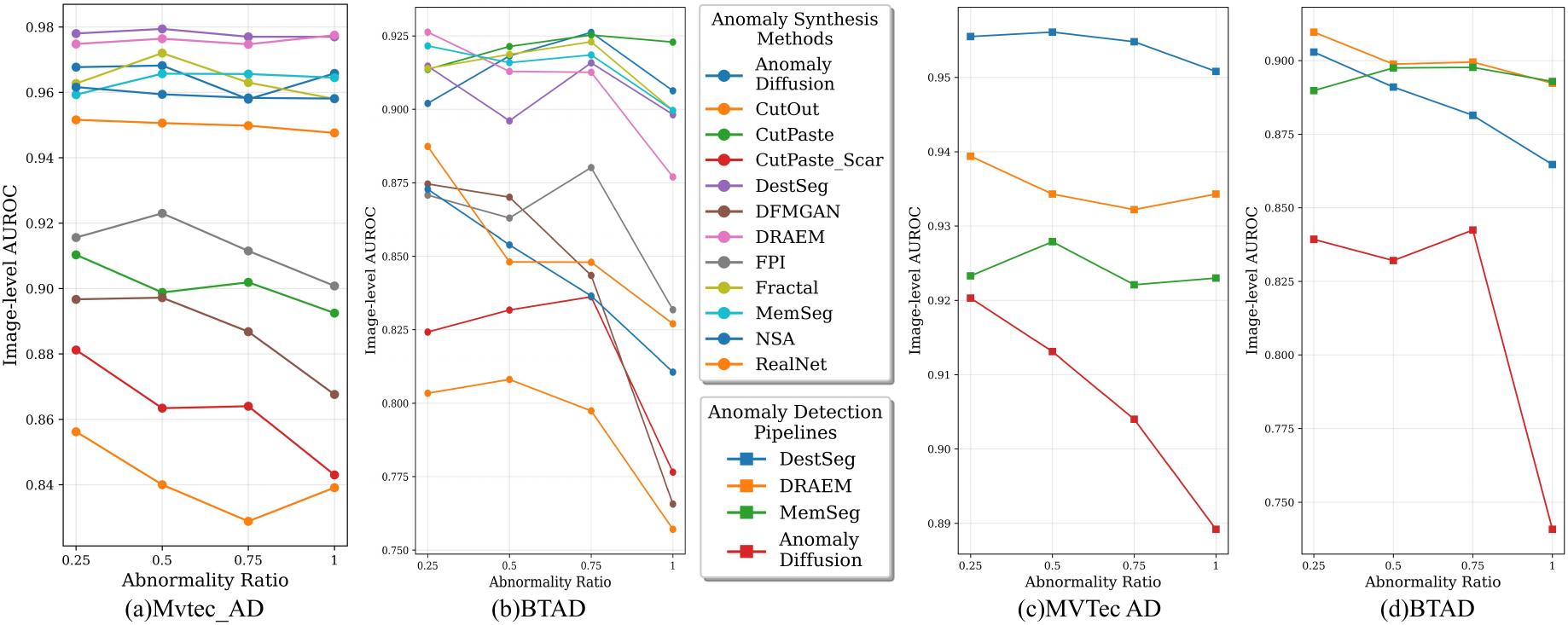}
\caption{Impact of anomaly ratio on anomaly synthesis methods performance: (a\&b) image-level AUROC of different anomaly synthesis methods on MVTec AD and BTAD. (c\&d) image-level AUROC of different anomaly detection pipelines on MVTec AD and BTAD. A performance degradation is often observed when the abnormality ratio is set to 1.}
\label{fig:dataset_performance_ratio}
\end{figure*}

\begin{figure*}[htbp]
\centering
\includegraphics[width=0.95\linewidth]{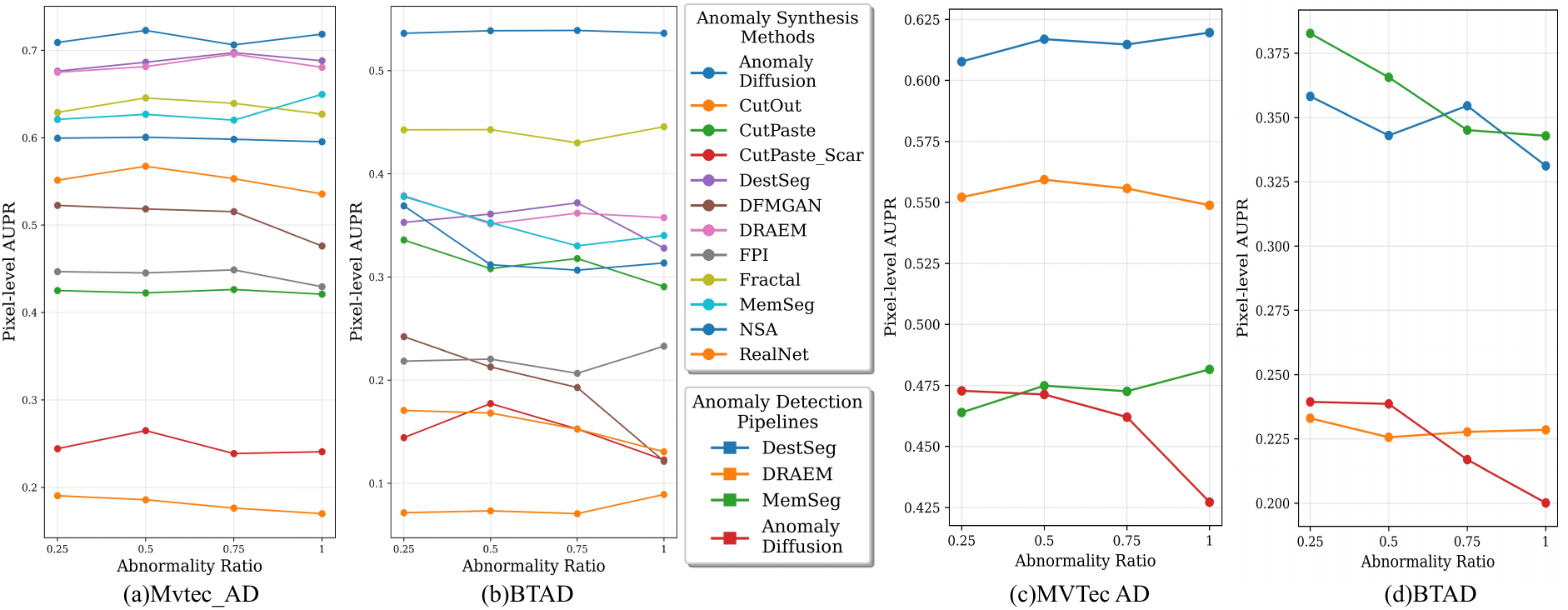}    
\caption{Impact of Anomaly Ratio on Anomaly Detection Pipelines Performance: (a\&b) pixel-level AUPR of different anomaly synthesis methods on MVTec AD and BTAD. (c\&d) pixel-level AUPR of different anomaly detection pipelines on MVTec AD and BTAD. A performance degradation is often observed when the abnormality ratio is set to 1.}
\label{fig:performance_ppl_ratio}
\end{figure*}

As shown in Table~\ref{tab:combined_scores_simple} and Fig.~\ref{fig:dataset_performance_ratio}, the experimental results under anomaly-free conditions demonstrate significantly inferior detection performance compared to scenarios containing anomalous samples. The significant difference observed demonstrate the efficacy of anomaly synthesis in enhancing detection capabilities across these algorithms. However, Fig.~\ref{fig:performance_ppl_ratio} illustrates that, after introducing synthetic anomalies, different ratios exhibit minimal performance variations, and no statistically significant optimal ratio point has been identified through current comparative analyzes. 
\begin{figure}[htbp]
    \centering
    \includegraphics[width=1.0\columnwidth]{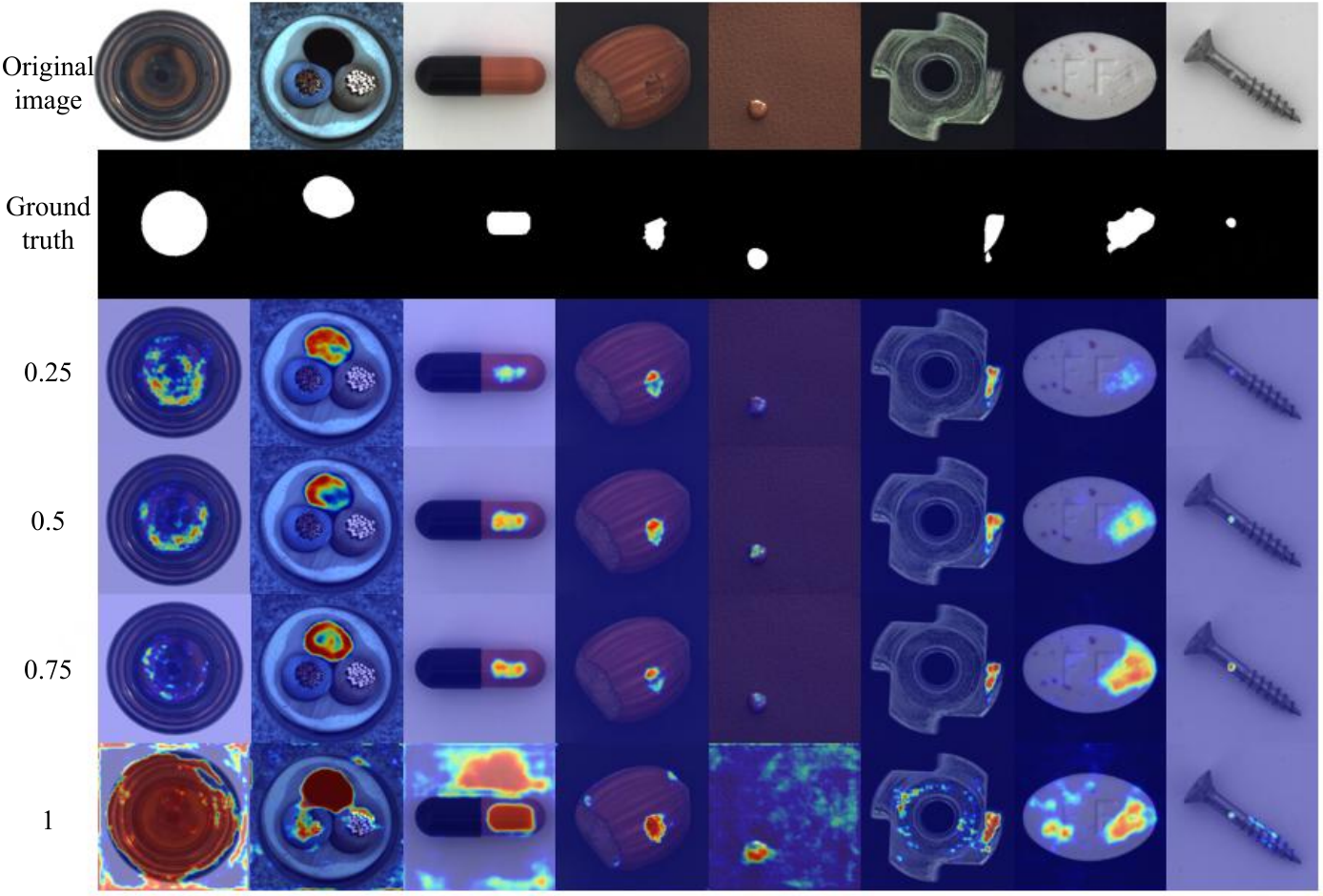}
    
    \caption{Visualization comparison of the synthesis method DFMGAN between different anomaly ratios.}
    \label{fig:vis_dfmgan}
\end{figure}
Notably, peak performance rarely occurs at higher anomaly concentrations. A particularly illustrative case is DFMGAN, which exhibits marked performance degradation at ratio=1.0. The visualization comparison is shown in Fig.~\ref{fig:vis_dfmgan}. This phenomenon stems from DFMGAN's design paradigm that leverages some real anomalies. Excessive synthetic anomalies during training induce overfitting to specific real anomaly patterns, consequently impairing generalization capacity on unseen anomaly types. In particular, According to the Fig.~\ref{fig:performance_ppl_ratio}, similar degradation is observed in AnomalyDiffusion detection algorithms at ratio = 1.0, likely attributable to their intrinsic anomaly-focused feature encoding mechanism that amplifies the risks of synthetic pattern overfitting.

For synthesis methods utilizing authentic anomalous samples, excessive incorporation of synthesized anomalous samples may increase the possibility of overfitting. In this context, while approaches such as DFMGAN and AnomalyDiffusion enhance the authenticity of generated anomalous samples, they do not exhibit superior performance in anomaly detection tasks. Although the generated anomalies more closely approximate the distribution of real-world anomalies, an overabundance of synthetic samples in the training set may cause the model to over-rely on specific features of these samples, thereby inducing overfitting and compromising the model's generalization capability on unseen data.

In particular, regarding the visual effect, we anticipate that the samples generated by AnomalyDiffusion will exhibit superior quality compared to DFMGAN, demonstrating greater authenticity and diversity.  AnomalyDiffusion achieves better anomaly detection performance despite its synthetic samples' higher fidelity. This observation implies that AnomalyDiffusion achieves a more effective equilibrium between authenticity and diversity in anomaly generation, thereby optimizing detection efficacy. In contrast, DFMGAN's anomaly synthesis process may overemphasize domain-specific anomalous features in certain scenarios, rendering detection models more susceptible to overfitting and impeding their ability to generalize across heterogeneous anomaly types.

Our investigation of ratio-dependent detection performance revealed that the seven top-performing methods exhibited greater robustness against ratio variations. Pearson's coefficients obtained from two independent datasets (-0.902409 and -0.701109) approached -1, demonstrating statistically significant strong negative linear relationships. These p-values (p = 0.000059 and 0.011074, respectively), being substantially below the 0.05 significance threshold, provide robust evidence that the observed correlations between ratio stability and detection efficacy are not attributable to random variation.

\begin{table*}[htbp]
\centering
\small 

\resizebox{\textwidth}{!}{%
\begin{tabular}{c|c|c|c|c|c|c|c|c|c|c|c|c}
\hline
{Method} & \makecell{Anomaly \\ Diffusion} & CutOut & CutPaste & \makecell{CutPaste \\ Scar} & DestSeg & DFMGAN & DRAEM & FPI & Fractal & MemSeg & NSA & RealNet \\
\hline
Mean & 0.9649 & 0.8410 & 0.9009 & 0.8629 & 0.9778 & 0.8871 & 0.9759 & 0.9127 & 0.9639 & 0.9638 & 0.9593 & 0.9499 \\
\hline
{Variance} & {2.27E-05} & {1.28E-04} & {5.44E-05} & {2.44E-04} & {1.25E-06} & {1.92E-04} & {1.84E-06} & {8.61E-05} & {3.45E-05} & {9.19E-06} & {2.63E-06} & {2.90E-06} \\
\hline
Max & 0.9682 & 0.8562 & 0.9103 & 0.8812 & 0.9794 & 0.8972 & 0.9775 & 0.9230 & 0.9720 & 0.9657 & 0.9616 & 0.9516 \\
\hline
Min & 0.9579 & 0.8288 & 0.8925 & 0.8430 & 0.9770 & 0.8676 & 0.9747 & 0.9008 & 0.9580 & 0.9593 & 0.9581 & 0.9476 \\
\hline
Range (Max - Min)& 0.0102 & 0.0274 & 0.0178 & 0.0382 & 0.0024 & 0.0296 & 0.0028 & 0.0223 & 0.0141 & 0.0064 & 0.0035 & 0.0040 \\
\hline
\end{tabular}%
}
\caption{Performance of anomaly synthesis methods on different proportions based on image AUROC and MVTec AD dataset.}
\label{tab:Performance_Proportion_MVTec}
\end{table*}

\begin{table*}[htbp]
\centering
\small 
\resizebox{\textwidth}{!}{%
\begin{tabular}{c|c|c|c|c|c|c|c|c|c|c|c|c}
\hline
{Method} & \makecell{Anomaly \\ Diffusion} & CutOut & CutPaste & \makecell{CutPaste \\ Scar} & DestSeg & DFMGAN & DRAEM & FPI & Fractal & MemSeg & NSA & RealNet \\
\hline
Mean & 0.9132 & 0.7915 & 0.9208 & 0.8172 & 0.9062 & 0.8385 & 0.9072 & 0.8614 & 0.9138 & 0.9139 & 0.8435 & 0.8526 \\
\hline
{Variance} & {1.22E-04} & {5.45E-04} & {2.56E-05} & {7.59E-04} & {1.10E-04} & {2.54E-03} & {4.46E-04} & {4.40E-04} & {1.03E-04} & {9.63E-05} & {6.99E-04} & {6.36E-04} \\
\hline
Max & 0.9262 & 0.8081 & 0.9253 & 0.8362 & 0.9158 & 0.8746 & 0.9263 & 0.8802 & 0.9230 & 0.9216 & 0.8728 & 0.8874 \\
\hline
Min & 0.9020 & 0.7571 & 0.9136 & 0.7765 & 0.8961 & 0.7657 & 0.8770 & 0.8318 & 0.8996 & 0.8996 & 0.8106 & 0.8270 \\
\hline
Range (Max - Min) & 0.0242 & 0.0510 & 0.0117 & 0.0597 & 0.0197 & 0.1089 & 0.0493 & 0.0484 & 0.0234 & 0.0220 & 0.0622 & 0.0604 \\
\hline
\end{tabular}%
}
\caption{Performance of anomaly synthesis methods on different proportions based on image AUROC and BTAD dataset.}
\label{tab:Performance_Proportion_BTAD}
\end{table*}

\subsubsection{Implications and Future Challenges}
To better utilize limited anomalous samples and prevent overfitting, technical improvements in anomaly synthesis methods are crucial. Excess synthetic samples fail not only to effectively enhance model performance but may induce overfitting to training data, resulting in suboptimal performance on unseen data. Therefore, it is essential to achieve an optimal equilibrium between the quantity and quality of generated samples while avoiding redundant or informationally redundant output. Adaptive generation strategies should be developed to more accurately simulate characteristics of scarce anomalous samples, enhancing diversity and authenticity through controlled stochastic processes. This optimization prevents overspecialization on specific anomaly features during generation, thereby improving model robustness and generalization capacity. 

\subsection{Correlation between Data Metrics and Performance}
\subsubsection{Comparative Experimental Framework}
Our comprehensive analysis across five datasets evaluates the correlation between detection performance and synthetic image quality metrics.   We compare generated anomalies with real defective samples using six established image generation metrics:   Structural Similarity Index (SSIM) for structural similarity, Peak Signal-to-Noise Ratio (PSNR) for pixel-level differences, Learned Perceptual Image Patch Similarity (LPIPS) for perceptual quality, Inception Score (IS) for diversity and quality assessment, with Fréchet Inception Distance (FID) and Kernel Inception Distance (KID) quantifying distributional discrepancies between synthetic and real data.   These metrics collectively provide a multidimensional evaluation of image generation fidelity, reconstruction accuracy, and perceptual characteristics.

\begin{figure}[htbp]
    \centering
    \includegraphics[width=0.9\linewidth]{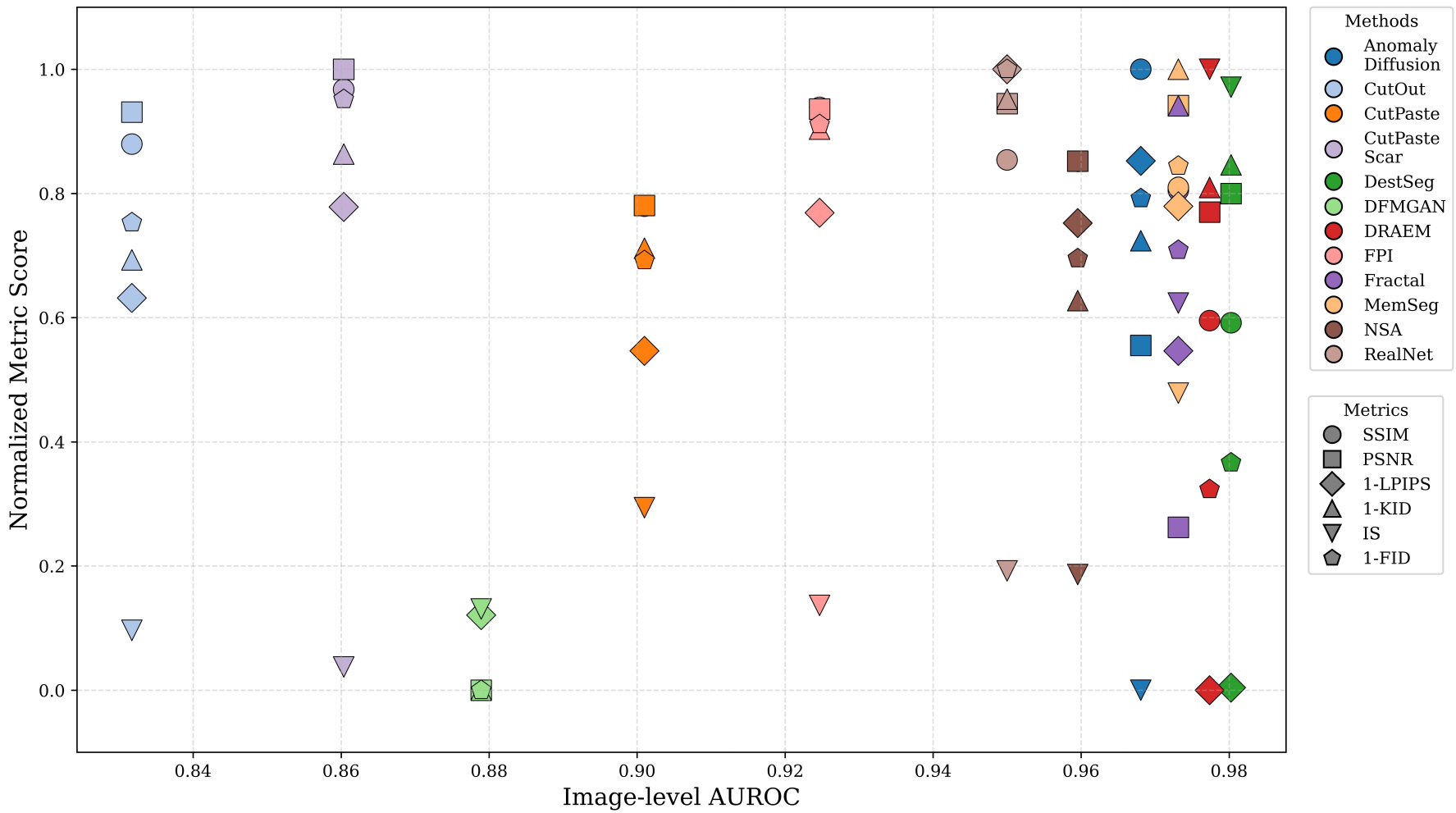}
    \caption{Correlation analysis between generated image quality metrics (SSIM, PSNR, LPIPS, IS, KID, FID) and anomaly detection performance (image-level AUROC) on MVTec-AD.}
    \label{fig:correlation_analysis}
\end{figure}

The analytical framework employs two-dimensional scatter plots, Fig.~\ref{fig:correlation_analysis} visualizing relationships between generation metrics and detection performance indicators, accompanied by Pearson correlation coefficients with statistical significance levels (p-values), as detailed in Table~\ref{tab:Correlation_AUCImage} . 

\begin{table*}
\centering
\resizebox{0.9\textwidth}{!}{%
\begin{tabular}{c|c|c|c|c|c|c|c|c|c|c}
\hline
Dataset & \multicolumn{2}{c|}{MVTec-AD} & \multicolumn{2}{c|}{BTAD} & \multicolumn{2}{c|}{VisA} & \multicolumn{2}{c|}{MPDD} & \multicolumn{2}{c}{MTD} \\
\hline
Comparison & Cor & P-value & Cor & P-value & Cor & P-value & Cor & P-value & Cor & P-value \\
\hline
\begin{tabular}{l}
SSIM vs AUROC
\end{tabular} & 0.0832 & 0.7972 & 0.1220 & 0.7056 & 0.1958 & 0.7056 & -0.1624 & 0.6141 & 0.2360 & 0.4603 \\
\hline
PSNR vs AUROC & -0.0472 & 0.8843 & 0.1314 & 0.6840 & 0.0336 & 0.6840 & -0.1813 & 0.5729 & -0.4960 & 0.1010 \\
\hline
LPIPS vs AUROC & 0.1068 & 0.7412 & -0.2138 & 0.5046 & 0.2278 & 0.5046 & 0.2053 & 0.5221 & -0.2194 & 0.4933 \\
\hline
IS vs AUROC & 0.6023 & 0.0382 & -0.0482 & 0.8817 & 0.6882 & 0.8817 & 0.5332 & 0.0742 & 0.0701 & 0.8286 \\
\hline
FID vs AUROC & 0.0358 & 0.9121 & -0.4318 & 0.1610 & -0.0163 & 0.1610 & 0.1645 & 0.6094 & -0.2800 & 0.3782 \\
\hline
KID vs AUROC & -0.4152 & 0.1796 & -0.3319 & 0.2919 & -0.3915 & 0.2919 & -0.0339 & 0.9167 & -0.3698 & 0.2368 \\
\hline
\end{tabular}
}
\caption{Correlation and P-values between six metrics and image-level AUROC across datasets.}
\label{tab:Correlation_AUCImage}
\end{table*}

\subsubsection{Analysis of Results}
The comprehensive analysis reveals weak correlations between anomaly detection performance metrics and image generation quality indicators. 
As evidenced by the tabular data and scatter plots of metric relationships, conventional image synthesis measures, including PSNR, SSIM, LPIPS, KID, FID, and IS demonstrate limited predictive capability for detection effectiveness.     

The computed Pearson coefficients and corresponding p-values predominantly exhibit two patterns: (1) Near-zero correlation coefficients coupled with statistically insignificant p-values , supporting the null hypothesis of no linear association;     (2) Moderately elevated correlation coefficients  accompanied by non-significant p-values, indicating insufficient statistical evidence for meaningful correlations.     

These dual patterns collectively confirm the absence of statistically robust linear relationships between synthetic image quality metrics and anomaly detection performance across the evaluated frameworks.
%
The weak correlation can be mainly attributed to two factors. Firstly, this discrepancy originates from the fundamentally divergent optimization objectives between conventional quality metrics and detection metrics. Image quality metrics predominantly assess reconstruction fidelity and visual similarity through pixel-wise comparisons and structural/textural analyzes. The metrics, typically computed at the image domain or holistic level, inadequately capture localized anomalies that characterize most synthetic anomalies. In contrast, detection-oriented metrics prioritize localized anomaly characteristics by design, systematically evaluating spatial precision to identify deviant patterns at the pixel or region level.

Secondly, existing generative models exhibit limitations in simulating the intrinsic complexity of real-world anomalies. When synthesized anomalies fail to precisely replicate the subtle morphological and contextual variations of authentic defects, significant discrepancies emerge between superficial image quality and practical detection utility. For example, certain synthetically generated anomalies may achieve high scores on traditional quality metrics while lacking discriminative features essential for effective detection, ultimately leading to suboptimal detection performance. 

\subsubsection{Implications and Future Challenges}

The limited predictive value for detection tasks of current evaluation requires the development of specialized assessment criteria for synthetic abnormal samples. An enhanced evaluation paradigm should concurrently operate at pixel-level and image-level dimensions, rather than rely solely on holistic image quality metrics. Such a dual-scale framework would enable more reliable quality assessment of synthetic anomalies while permitting prediction when utilizing data for detector training, thereby conserving human and material resources typically expended in repetitive experimental iterations of detection tasks.

This challenge parallels difficulties encountered in evaluation of natural image generation. Some methods assess generation quality through combined global-local analyzes. For example, portrait synthesis validation examines both overall reasonability and localized correctness. Drawing inspiration from these approaches, synthetic anomaly evaluation should incorporate Global consistency and Local fidelity. The establishment of such multidimensional evaluation standards represents a critical research direction for improving synthetic anomaly generation and application efficacy.


\subsection{Performance of Mixed Synthesis methods}

\subsubsection{Analysis of Results}
\begin{figure}[!t]
    \centering
    \includegraphics[width=1.0\columnwidth]{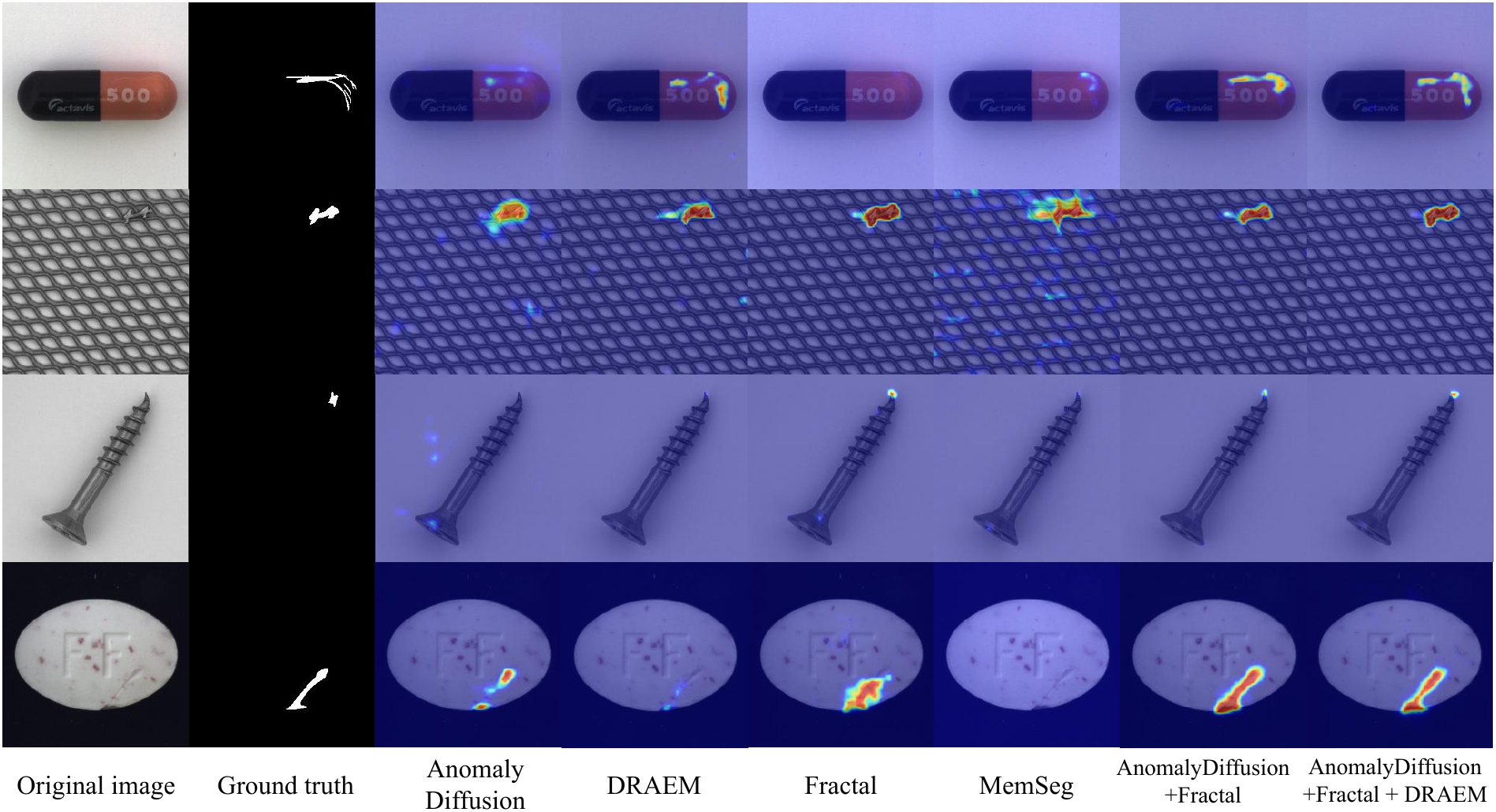}
    
    \caption{Visualization comparison between single synthesis method and mixed synthesis method.}
    \label{fig:vis_mutli}
\end{figure}

\subsubsection{Comparative Experimental Framework}
\begin{figure}[t]
    \centering
    \includegraphics[width=0.85\linewidth]{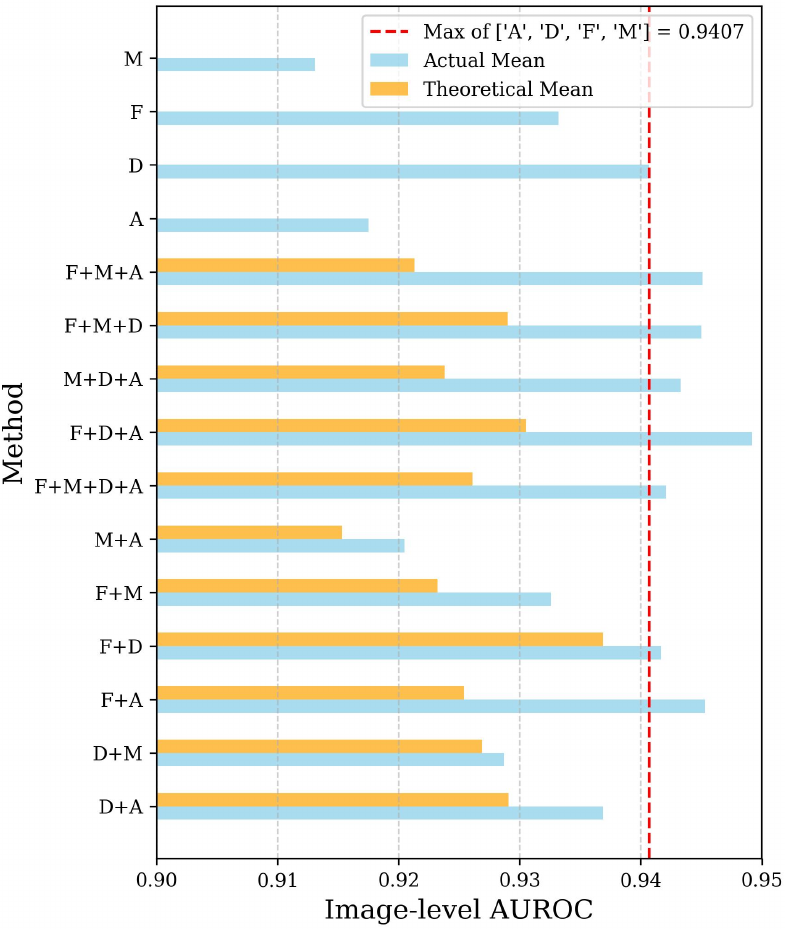}
    \caption{
        Comparison of performance (image-level AUROC) using mixed synthesis augmentation methods versus individual approaches. 
        \textbf{Blue part}: Actual performance of hybrid synthesis methods.
        \textbf{Yellow part}: Theoretical performance computed from individual methods' results.
        \textbf{Red line}: Maximum AUROC achieved by any single synthesis method.
    }
    \label{fig:performance}
\end{figure}

This section is based on four synthesis algorithms—Fractal, DRAEM, AnomalyDiffusion, and MemSeg—to construct and evaluate eleven hybrid combinations through training and testing, aiming to assess their synergistic effects in anomaly synthesis tasks. The experiments use the weighted average of performance metrics across the dataset as a reference. The results of the hybrid methods are compared with those of each individual method and theoretical combined means. The comparison of visualization is shown in Fig.~\ref{fig:vis_mutli}. And bar charts are plotted for different detection pipelines, as shown in Fig.~\ref{fig:performance}, to analyze the performance differences among various combinations.
From Fig.~\ref{fig:performance}, it can be observed that most hybrid methods outperform the best results achieved by individual methods. A few hybrid methods show slightly lower performance than the optimal single method, but still surpass the weaker individual methods. Additionally, by comparing the theoretical and actual performance of hybrid methods, we find that the actual performance of hybrid methods generally exceeds the theoretical mean, indicating that the combination of multiple methods indeed produces synergistic effects.

This improvement primarily stems from the diversity, complementarity, and robustness of the different algorithms. Each algorithm has distinct strengths and characteristics when handling anomalies. Combining these methods can cover a broader range of anomaly patterns, thereby enhancing overall detection performance. Furthermore, hybrid methods, by integrating the results of multiple algorithms, reduce the sensitivity of individual methods to specific noise or data distributions, thus improving overall robustness.

Notably, mixed methods incorporating AnomalyDiffusion show significant improvement in the pixel-level metric compared to others. AnomalyDiffusion generates abnormal samples with high quality and diversity at pixel-level, better simulating real-world anomaly data. This allows the model to learn more refined anomaly features during training, while other methods primarily focus on image-level.

\subsubsection{Implications and Future Challenges}
Although the diversity, complementarity, and robustness of different methods are advantages of hybrid approaches, selecting optimal combination of methods and adjusting their weights remains a complex optimization problem.
The significant improvement in pixel-level performance when AnomalyDiffusion is incorporated highlights necessity of including high-quality pixel-level samples. In other words, it is worth exploring anomaly synthesis methods that operate at both image and pixel level.

\section{Conclusion} 
We introduce ASBench in this paper, the first comprehensive benchmark for image anomaly synthesis, featuring 12 anomaly synthesis methods, 4 anomaly detection pipelines, and 5 industrial datasets across 4 key evaluation dimensions. Through extensive experiments, we have gained crucial insights into the performance of anomaly synthesis, including the discovery that no single method is universally optimal; the non-linear influence of synthetic data ratios, where higher proportions do not guarantee better performance; the weak correlation between conventional image quality metrics and downstream detection efficacy; and the notable performance boost from hybridizing complementary synthesis methods. On top of these findings, we present several intriguing future lines for image anomaly synthesis. For example, future work could focus on (1) designing anomaly synthesis methods with greater generalizability to balance diversity and realism; (2) developing adaptive synthesis strategies to optimize the
trade-off between the quantity and quality of synthetic samples; (3) constructing novel task-oriented
evaluation metrics that can reliably predict detection performance; and (4) systematically optimizing
the combination and weighting mechanisms when hybridizing multiple synthesis approaches. 

\bibliographystyle{plain}
\bibliography{egbib}

\end{document}